# Towards Real-time Drowsiness Detection for Elderly Care

Boris Bačić
*Engineering, Computer and Mathematical Sciences*
*Auckland University of Technology*
Auckland, New Zealand
0000-0003-0305-4322

Jason Zhang
*Engineering, Computer and Mathematical Sciences*
*Auckland University of Technology*
Auckland, New Zealand
grr6525@autuni.ac.nz

*Abstract*—The primary focus of this paper is to produce a proof of concept for extracting drowsiness information from videos to help elderly living on their own. To quantify yawning, eyelid and head movement over time, we extracted 3000 images from captured videos for training and testing of deep learning models integrated with OpenCV library. The achieved classification accuracy for eyelid and mouth open/close status were between 94.3%-97.2%. Visual inspection of head movement from videos with generated 3D coordinate overlays, indicated clear spatiotemporal patterns in collected data (yaw, roll and pitch). Extraction methodology of the drowsiness information as timeseries is applicable to other contexts including support for prior work in privacy-preserving augmented coaching, sport rehabilitation, and integration with big data platform in healthcare.

## I. Introduction

With aging, sleep quality is often associated with acute or chronic pain, discomfort, irregular sleeping patterns, daytime naps, fatigue and health degradation. It is also getting easier to choose to sit in front of the TV (or computer) screen than to move and spend time while enjoying outdoor lights. Taking a short nap after lunch may be common in many cultures and help with alertness and energy levels. However, it is also known that for people experiencing poor sleep quality, daily naps would rather worsen than alleviate such problems. Having access to drowsiness information collected from the environment(s) in which we spend most of our daytime could provide opportunities for improving overall life quality for a growing number of people, who may not even be aware of how often and for how long their daily napping pattern has changed.

To develop technology to enhance improving life quality for people experiencing sleeping problems, including elderly living on their own we need to ask the following research questions:

1. Can we extract drowsiness information from videos while preserving privacy using consumer-grade technology?
2. If eyelid position is not always visible e.g. for people who wear darkened glasses or a hat, what other visual downiness cues could be extracted?
3. In addition to traffic safety, what other research areas may be recommended for a broader application of drowsiness data and presented methodology?

## II. Background: Related Contexts and Prior Work

The United Nations organisation projected that by 2030, people over 60 years of age will account for 16% of the world population. And by 2050, one in six people will be over 65 years old [1]. Compared to living in nursing homes providing additional medical care, a substantial proportion of people prefer to live in their own homes [2]. The increase in aeging population and shortage of nursing staff is raising concerns about the costs and affordability of heathcare services. However with technology development, we expect to automate seome aspects of helathcare service which may be available on a 24x7 basis. Sleep monitoring technology is already becoming available to help people living on their own [3]. In general healthcare monitoring systems can fall into two categories using data from: (1) wearable devices and (2) fixed monitoring equpmen. In addition to wearable devices, today smartwatches can also cover some data colletion related to sleep quality. Sleep monitoring devices would typaclly collect breathng rates, Electroencephalography (EEG), and hearbeat data [4]. However, in reality, wearable devices may not confrotable to wear on a 24x7 bases. As a less obtrusive alternative, is to install fixed monitoring equipment [5-7], which for drowinsess detection could use computer vision focusing on face area, eyes, yawning and sudden nodding of the head followed by rapid head raising.

### A. Face Detection

Before the advent of deep learning applications in computer vision, a popular method for face detection was the Viola-Jones algorithm [8]. Since its' first publication in 2001, Viola-Jones algorithm is still in use with some incremental improvements such as robustness and redundancy processing [9] optimisation. The growing popularity and success of deep learning approaches, is also applicable to face detection research area. For example, H. Li et al. proposed a method for face detection by using cascaded CNN [10]. The use of Multilayer Convolutional Neural Network (MTCNN) was extended to face detection and facial expression recognition [11]. Compared with the algorithm proposed in 2017, MTCNN reduces the number of convolution kernels, increases the depth of the network, and improves the efficiency of image processing [11-13]. MTCNN contains three cascaded multi-task convolutional neural networks, including Proposal Network (P-Net), Refine Network (R-Net), and Output Network (O-Net). MTCNN integrates face area detection and face keys point detection, and given its reported performance, it can run smoothly on the mobile phone systems or embedded devices [12].

## B. Head Pose Estimation

Pose estimation from 2D videos is still considered a challenge. In 2016, S. Malik [14] uses a set of six defined facial key points of a single picture with four distance coefficients, and calibrated camera parameters to obtain 3D head pose estimation data using open-source OpenCV library (https://opencv.org). In 2019, Oprea et al. proposed a real-time head pose estimation algorithm also relying on open-source OpenCV library, advancing the state-of-the-art by using a time-of-flight depth camera and with the same six facial landmark points [15]. As an alternative, Ruiz et al. proposed a head pose estimation algorithm based on trained multi-loss convolutional neural network working without the need for direct transformation of provided facial key points [16].

## C. Drowsiness Detection

Drowsiness detection has a strong presence in driving safety monitoring. In computer vision, drowsiness detection based on facial features can be regarded as a two-label classification problem. For example, Bhaskar designed a helmet-like wearable device that can monitor the driver's blinking frequency, heartbeat rate, breathing rate and head movement, for enhanced driver's drawsingess detection [17]. The accuracy of Bhaskar's wearable device can reach 70%. However, wearing the device for a long time will increase the driver's fatigue and additional equipment is likely to obstruct the driver's line of sight, arguably defeating the purpose of intended driving safety. Other examples include combination of traditional computer vision approaches such as Histogram of Oriented Gradients (HOG) combined with other machine learning techniques [18]. In Revelo et al. approach [19], for example, infrared real-time video processing starts with the Viola-Jones algorithm [8] for measured head inclination angles of up to $21°$. After detected face region, the subsequent region of interest (ROI) surrounding one eye was identified based known face proportions. The eye ROI is cropped, further normalized and transformed as a binary image data input for traditional classifier such as Multi-Layer Perceptron (MLP) achieving 84% accuracy.

Many researchers use Convolution Neural Networks (CNN) for drowsiness detection. Jabbar at. al. proposed a drowsiness detection system based on deep learning technology deployed in Android applications reporting achieved accuracy of over 80% [20]. By compressing the model and redesigning the neural network structure, Reddy et al. [13] proposed a detection system that can be deployed in embedded devices. The detection system capable of processing open and close states of the eyes and mouth, achieved accuracy of 89.5%

## D. Prior Work

Previous work includes silhouette-based video filtering, human motion modelling and analysis (HMMA) and big data heath care platform design and evaluation [21-23]. The data format associated with drowsiness are aligned with privacy-preserving initiative, where to preserve diagnostic information, we produced a pseudo 2D and 3D silhouette variations with golf club tracking. Both silhouette variations can show limb separation from other parts of the body for chosen camera views evaluated on a golf driving range [24, 25]. In our experimental work, we used a range of video cameras (mobile, web cams, camcorders, high speed action cameras, and Kinect). For Kinect system, we experimented with depth and infrared vision that are equally applicable for coaching and fall detection in elderly living environment settings with a degree of privacy protection. For golf coaching data collection research, head position allowing "eye on the ball" is one of the critical coaching aspects associated with stance and swing "through the ball" where it is expected from players to not move their head to early into follow-through phase of the movement [22, 26, 27]. In another case study on an elite-level tennis player returning to sport after shoulder surgery, we used a high-speed camera mounted above the player that could capture ball toss during the serve and record her head and eye position through the impact with the ball, allowing analysis and recommending intervention from collected video evidence [28, 29]. Another opportunity to use data in healthcare is to integrate drowsiness real-time data streaming with next-generation of analytical healthcare platforms supporting big data technology and horizontal scaling (expanding the system using low-cost commodity hardware), which for data protection are to be installed on premises, with dedicated communication lines and operating within the national borders [21].

Obtaining visual annotations and privacy-preserving timeseries extracting eyelid, moth and head movements are important to advance prior work as secondary objective of the study.

## III. METHODOLOGY

For drowsiness monitoring and video data extraction flowchart, there are three key processing steps (Fig. 1).

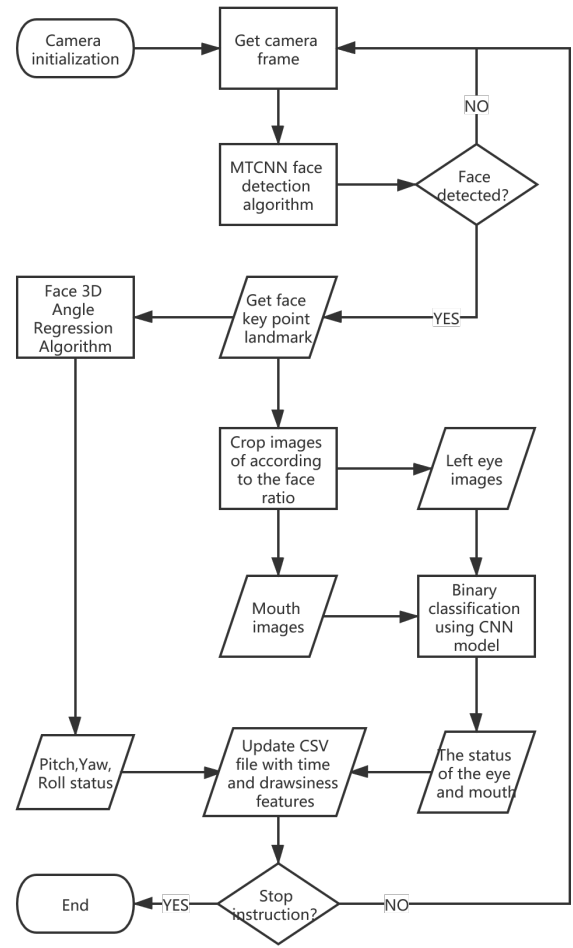

Fig. 1. Drowsiness monitoring and video data extraction flowchart.

For our algorithm (presented as flowchart in Fig. 1), the first step is face detection, i.e. detecting whether there is someone in the current frame, followed-up with the head and face detection. If there is no face detected, the system will repeat this operation on the next frame without completing the second step. In this step, we use MTCNN, not only because of the high detection accuracy and processing speed, but also because MTCNN can obtain the boundary coordinates of the face and the coordinates of five key points including eyes, mouth, and nose [30].

The next step is to identify the key points associated with facial landmarks (left eye and mouth as regions of interest). In addition, the algorithm (Fig. 1) also process the facial key points, face boundaries and the face ratio (Fig. 2) detected by MTCNN [30]. We trained a selection of CNN algorithms (Table 1) to perform two classification tasks associated with left eye and mouth open or close status.

The third step is to obtain the head pose estimation through solvePnP(·) function, which is included in the OpenCV libraries.

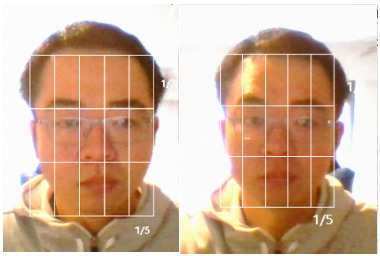

Fig. 2. The face proportions are constant for camera distance variations

### A. Classification Methods for Eyes and Mouth Regions

For algorithm efficiency, as to increase the processing speed, while reducing computational costs, we only capture the images of the left eye and mouth. For reported personalized model development, we manually extracted 3000 images showing balanced distribution of open and closed eyes and mouth status. The location of the key points, including eyes, mouth, and nose are provided by MTCNN as real-time ROI detector. From estimated face proportions (Fig. 2), the eyes account for about 20% of the overall width of the face, and the height of the eye accounts for about 15% of the face height. Therefore, the coordinates of the upper left and lower right corners of the eyes can be obtained from the eye key points. Similarly, the width of the mouth accounts for about 30% of the face width, and the height accounts for about 15% of the overall height, in agreement with the 2019's study on facial analysis measurements by golden proportion [31].

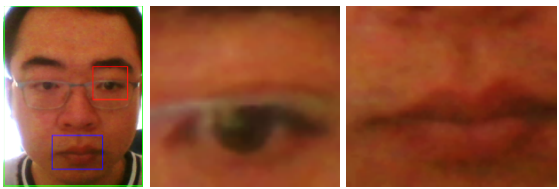

Fig. 3. Cropped eye and mouth region of interests.

A selection of the eye and mouth images cropped to scale that were used for model training are shown in Fig. 3 and Fig.

4. For modelling purposes, we divided the collected dataset into training (90%) and testing (10%) portions.

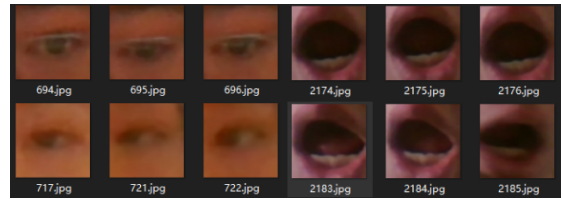

Fig. 4. Training dataset example.

### B. Head pose estimation

The objective of head pose estimation task is to estimate head position in virtual 3D space from 2D images and generate yaw, pitch and roll values. To facilitate 2D to 3D face estimation relying on estimated face landmarks, we combine MTCNN model with solvePnP(·) function from OpenCV library.

Given our intention to look for spontaneous relaxation of neck muscles and characteristic head movement patterns associated with drowsiness such as those depicted in Fig. 8, we obtained five key points using MTCNN and calculated additional key point representing the chin centre as face bottom landmark (Fig. 5). Using calculated chin centre is not susceptible to temporary chin occlusions such as habitual hand movement during the yawn. For enhanced conversion accuracy, we used solvePnP(·) function, which uses produced six facial landmarks as 2D image points (Fig. 5) and returns estimated head pose as 3D model points (Fig. 8).

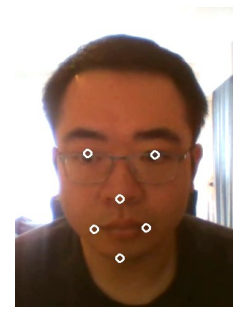

Fig. 5. The five key points obtained by MTCNN plus the calculated key point of the chin.

Finally, the relevant data about the eyes, mouth and estimated 3D head movement are saved as extracted timeseries for further research and analysis.

## IV. EXPERIMENTAL RESULTS

For personalised model performance evaluation and benchmarking purposes, in our experiments, we compared different candidate CNNs models such as: Alexnet, Vggnet, Googlenet, Resnet and the Mobilenet. The operating system of the notebook used in the experiment is Windows10 18363.1.16. The CPU is Intel Core I5-4210M, and the GPU model is Nvidia GEFORCE 940M.

The accuracy and loss of each CNN training are shown in Fig. 6.

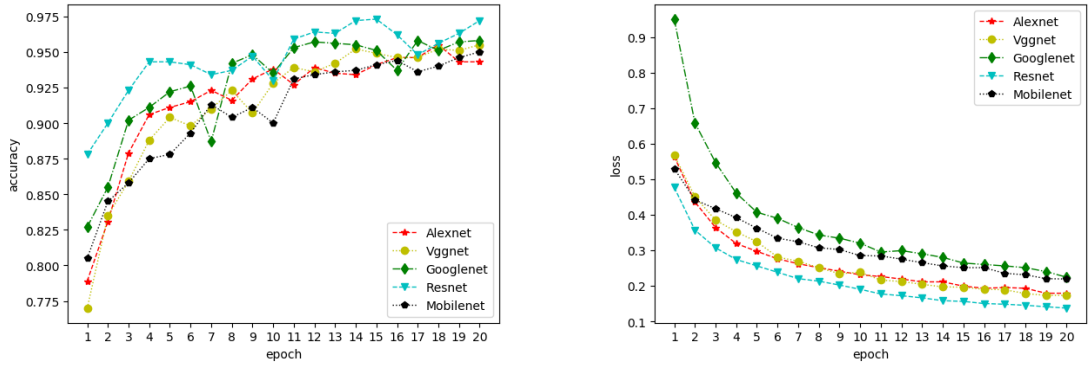

Fig. 6. Accuracy and loss comparisons for tested models.

Video input (Fig. 7) and produced output video overlays and timeseries data are shown in Fig. 8 and Fig 9.

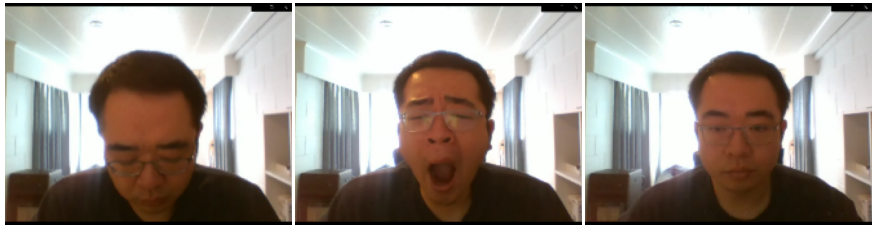

Fig 7. Sample of frame screenshots captured videos.

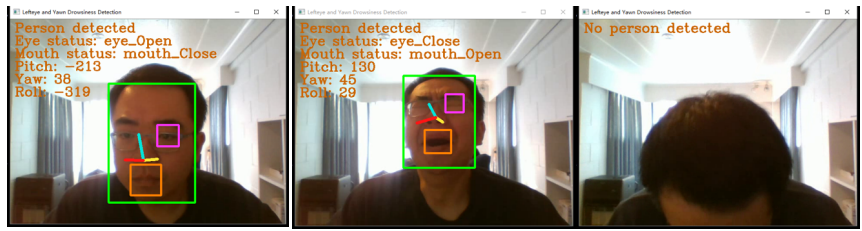

Fig. 8. Sample of screenshots with video overlays from developed experimental system.

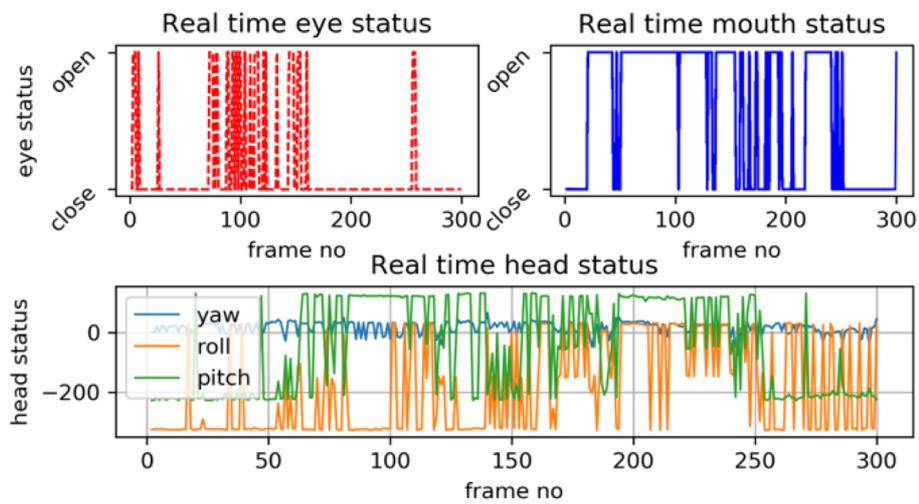

Fig. 9. Part visual data captured by experimental videos.

From results comparisons (Fig. 6), by the 20th cycle, the loss tends to be stable, and the accuracy rate does not fluctuate greatly. The results are summarised in Table 1 and Fig. 10.

TABLE I. SUMMARY OF ACCURACY AND LOSS COMPARISONS

| CNN Model | Accuracy | Loss |
|---|---|---|
| Alexnet | 0.943 | 0.179 |
| Vggnet | 0.955 | 0.173 |
| Googlenet | 0.958 | 0.224 |
| **Resnet** | **0.972** | **0.137** |
| Mobilenet | 0.950 | 0.219 |

Figure 7 shows the precision and recall factors for tested CNN models that were derived from confusion matrix reports.

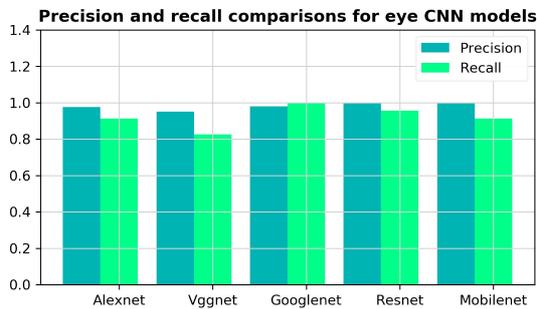

Fig. 10. Precision and recall comparisons for CNN models.

As expected, all CNN models show a similar performance suggesting that Resnet and Googlenet demonstrated the best accuracy and performance.

To validate intermediate processing results of head pose estimation, we calculated the 3D coordinate system following the head moment with its origin between the eyes and generated video overlays (Fig 8). Annotated videos showing head, mouth and eye movements were visually inspected as intermediate processing results and compared with generated timeseries capturing eye, mouth and 3D head pose status in each frame (Fig. 8).

## V. DISCUSSION

Given the proof-of-concept nature of our experiments, we should disclose the following limitations: we did not use a camera with zoom and infrared functions, so in the case of insufficient light, the accuracy rate may drop significantly. Another limitation is our 3D head pose estimation. If the subject's head tilts forward too much during a nap and it is too close to the camera (Fig. 8), the MTCNN model cannot detect the face of the user, which may cause the head pose estimation to fail temporarily. For such cases, we may supply a copy of the last valid data until the face is detected again. The third limitation is that reported data correspond to a personalised model, which is expected to have higher accuracy than a model for a similar group of people or global population sample. We also expect that for other personalised models some CNN may perform better than others, which was the case reported in this study. Within the scope of the study, for 3D head position estimation, we did not adjust camera calibration parameters for the camera used nor did we try an external validity study with sub-millimetre precision 3D motion capture data system (which could be a separate study). Based on the visual inspection of head movement data patterns and video overlays, we found, therefore, that the presented method is of acceptable quality (albeit with a degree of systematic error depending on the camera) and can be trained for real-life habitual gesture of partial mouth covering during yawning. Another incremental research opportunity to pursue includes repurposing mobiles, tablets and other near-obsolete low-power devices, so that our methodology implementing more compact CNN models would signify digital inclusion of a broader population aspiring to more active and fulfilling life.

The proposed joint use of MTCNN and other CNNs to detect sleepiness in the daily life of elderly people living alone and to generate corresponding raw data and statistical information has promising potential to assist medical staff in preventive medicine, rehabilitation, early diagnostics and in determining ongoing medical issues we associate today with modern precision medicine.

## VI. CONCLUSION

To provide support to people experiencing sleeping problems including the elderly living on their own we developed solutions to extract drowsiness information from video. The generated corresponding statistical data may assist healthcare professionals as well as friends and family to investigate changes in daily napping habits that may be linked to fatigue, discomfort, causes of chronic pain, sleeping disorders manifestation and other symptoms of health degradation.

We demonstrated that our proof of concept, MTCNN, 2D to 3D pose estimation combined with other CNNs to detect sleepiness in the daily life can run locally on low-cost consumer-grade computing platforms intended to preserve privacy by not transmitting the video stream but rather producing text files containing timeseries data. In addition to eyelid open or close status, we also recorded yawning information from detected mouth movements. Projected head tilt in 3D from 2D video frame can provide additional information about the observed subject's attention focus, sleep position and transitions such as from sitting into napping head movement pattern from the relaxation of neck muscles.

Future work will consider integration of infrared and additional multi-camera sources, wearables and low-cost hardware as the IoT processing infrastructure for local data exchanged via batch mode or streaming with near-future big data cloud processing platforms. As a secondary objective and follow up to this research, we intend to enable methodology transfer to numerous related contexts including rehabilitation and augmented coaching systems and technology (ACST), and integration with near-future big data analytical platforms in healthcare that will allow data streaming from wearables and various activity monitoring devices based on the presented proof of concept.

ACKNOWLEDGEMENT

We wish to acknowledge contributors to the open source software community including OpenCV documentation and libraries, which were important success factors for us to complete this research.